\documentclass[runningheads]{llncs}
\usepackage[T1]{fontenc}
\usepackage{graphicx,verbatim}
\usepackage{amsmath}
\usepackage{multirow}
\usepackage{amssymb}
\usepackage{bm}
\usepackage{ulem}
\usepackage[colorlinks]{hyperref}
\usepackage{makecell}

\usepackage{cleveref}
\usepackage{booktabs}
\usepackage{marvosym}

\begin{document}
\title{Dyna3DGR: 4D Cardiac Motion Tracking with Dynamic 3D Gaussian Representation}
\titlerunning{Dyna3DGR}
%

\authorrunning{X. Fu et al.}

\author{Xueming Fu\inst{1,2} 
\and Pei Wu \inst{1,2} 
\and Yingtai Li\inst{1,2} 
\and Xin Luo\inst{1,2} 
\and Zihang Jiang\inst{1,2,3} 
\and Junhao Mei\inst{5} 
\and Jian Lu\inst{5} 
\and Gao-Jun Teng\inst{5} 
\and S. Kevin Zhou\inst{1,2,3,4} 
$^{\href{mailto:skevinzhou@ustc.edu.cn}{\textrm{\Letter}}}$}

\institute{School of Biomedical Engineering, Division of Life Sciences and Medicine, University of Science and Technology of China (USTC), Hefei Anhui, 230026, China \and
Center for Medical Imaging, Robotics, Analytic Computing \& Learning (MIRACLE), Suzhou Institute for Advance Research, USTC, Suzhou Jiangsu, 215123, China \and
Jiangsu Provincial Key Laboratory of Multimodal Digital Twin Technology, Suzhou Jiangsu, 215123, China
\and
State Key Laboratory of Precision and Intelligent Chemistry, USTC, Hefei Anhui, 230026, China 
\and
Center of Interventional Radiology \& Vascular Surgery, Department of Radiology, Medical School, Zhongda Hospital,  Southeast University, Nanjing 210009, China 
}

\maketitle              
\begin{abstract}
Accurate analysis of cardiac motion is crucial for evaluating cardiac function. While dynamic cardiac magnetic resonance imaging (CMR) can capture detailed tissue motion throughout the cardiac cycle, the fine-grained 4D cardiac motion tracking remains challenging due to the homogeneous nature of myocardial tissue and the lack of distinctive features. Existing approaches can be broadly categorized into image-based and representation-based, each with its limitations. Image-based methods, including both traditional and deep learning-based registration approaches, either struggle with topological consistency or rely heavily on extensive training data. 
Representation-based methods, while promising, often suffer from loss of image-level details.  
To address these limitations, we propose {\bf Dyna}mic {\bf 3D} {\bf G}aussian {\bf R}epresentation ({\bf Dyna3DGR}), a novel framework that combines explicit 3D Gaussian representation with implicit neural motion field modeling. Our method simultaneously optimizes cardiac structure and motion in a self-supervised manner, eliminating the need for extensive training data or point-to-point correspondences. Through differentiable volumetric rendering, Dyna3DGR efficiently bridges continuous motion representation with image-space alignment while preserving both topological and temporal consistency. Comprehensive evaluations on the ACDC dataset demonstrate that our approach surpasses state-of-the-art deep learning-based diffeomorphic registration methods in tracking accuracy. The code will be available in https://github.com/windrise/Dyna3DGR.


\end{abstract}
\section{Introduction}

Accurate estimation of myocardial motion is essential for evaluating cardiac function and diagnosing myocardial diseases\cite{Myo2017Infarction}.
Dynamic cardiac motion reconstruction provides comprehensive spatiotemporal information throughout cardiac cycle, enabling clinicians to better analyze physiological cardiac dynamics for improved diagnostic accuracy and treatment planning.
Tagged cardiac magnetic resonance imaging (t-CMR)\cite{amzulescu2019myocardial} serves as the gold standard for assessing myocardial motion using intrinsic markers\cite{ye2021deeptag}. However, due to its complex acquisition process, recent research has increasingly focused on estimating motion from untagged CMR images\cite{zheng2019explainable,morales2019implementation,krebs2019learning,yu2020foal_cvpr}.
Technically, cardiac motion estimation approaches can be broadly categorized into two main streams: image-based and representation-based.
In the image-based methods, researchers have developed various non-parametric registration methods that rely on mathematical priors and optimization techniques. These include incorporating free-form deformations with B-splines\cite{rueckert1999nonrigid-FFD}, optical flow\cite{becciu2009multi}, and biomechanics-informed\cite{qin2020biomechanics} approaches to achieve accurate correspondence mapping. 
While these methods have shown promise, they often struggle with preserving topological consistency during deformation. 
To address this limitation, 
diffeomorphic registration methods\cite{beg2005computing,shen2019region,avants2008symmetric} have introduced topology-preserving constraints.
However, these non-parametric registration approaches remain computationally intensive and sensitive to image noise.
%
The advent of deep learning has revolutionized image-based cardiac motion estimation. Data-driven deep registration approaches have demonstrated superior performance in preserving topological consistency and maintaining long-term temporal coherence compared to traditional registration methods \cite{balakrishnan2019voxelmorph,kim2022diffusemorph,qin2023fsdiffreg,chen2023transmatch,meng2024correlation,yang2025bidirectional}. 
However, their effectiveness is inherently constrained by the availability of extensive training data, and they often face challenges in generalizing across datasets with different distributions.

While untagged CMR provides clear visualization of cardiac structures that can be precisely segmented, the inherent elasticity and homogeneous nature of myocardial tissue present significant challenges for accurate motion tracking in image space due to the lack of reliable natural landmarks within the tissue.
To alleviate this problem, another line of research explores cardiac motion estimation in alternative representation spaces. Guo {\it et al.} \cite{guo2021unsupervised} have proposed an unsupervised approach to extract stable landmarks from volumetric images, using optimal transport theory with topological constraints for motion field estimation. However, this approach can be sensitive to noise and may not achieve sufficient precision. Meng {\it et al.} \cite{meng2023deepmesh} have adopted fixed-vertex mesh representations with template topology ($\sim$20,000 vertices) as stable identifiers across different subjects and cardiac cycles, and estimate vertex motion from six different view sequences to reconstruct myocardial deformation. While effective, this approach may lose fine-grained image details. Yuan {\it et al.} \cite{yuan20234d} have explored using implicit neural representations through signed distance field to model the myocardium, enabling continuous shape representation. However, their approach focuses primarily on global shape deformation modeling, making it challenging to capture fine-grained local motion details.
While these representation-based approaches show promise in breaking through the performance ceiling of image-based methods, there remains a critical need for a unified framework that can both accurately represent cardiac anatomy and seamlessly bridge the gap between representation space and image space.
To tackle the challenges, we propose {\bf Dyna}mic {\bf 3D G}aussian {\bf R}epresentation ({\bf Dyna3DGR}), a novel framework that combines explicit 3D Gaussian representation with implicit neural motion field modeling. Our approach simultaneously optimizes cardiac structure and motion reconstruction in a self-supervised manner, eliminating the need for extensive training data or dense correspondences across cardiac cycles. Through differentiable volumetric rendering, Dyna3DGR efficiently bridges the gap between 3D Gaussian representation and image-space. Our key contributions can be summarized as follows:
\begin{enumerate}
\item We propose a self-supervised optimization framework for 4D cardiac motion estimation that simultaneously optimizes cardiac structure and motion estimation, eliminating the dependency on extensive training data that is commonly required by existing image-space methods.
\item Through the unique integration of explicit 3D Gaussian representation and implicit neural deformation field modeling in Dyna3DGR, it effectively fills the gap between representation space and image space. This hybrid design not only preserves topological consistency but also achieves accurate motion tracking without requiring explicit dense correspondence, addressing the shortcomings of existing representation-based approaches.
\item Comprehensive evaluations on the ACDC dataset demonstrate that our approach surpasses state-of-the-art diffeomorphic registration methods in tracking accuracy, validating the effectiveness of our proposed framework.
\end{enumerate}

\begin{table}[t]
    \centering
    \caption{Comparison of different methods.}
    \label{tab:comparison}
    \begin{tabular}{l|l|c|c}
    \toprule
    Methods & Representation & Image Details & No Extra Data \\
    \midrule
     Traditional registration \cite{rueckert1999nonrigid-FFD,becciu2009multi,qin2020biomechanics}& Pixel/Voxel & \checkmark & \checkmark \\
DL-based registration \cite{balakrishnan2019voxelmorph,kim2022diffusemorph,qin2023fsdiffreg,chen2023transmatch,meng2024correlation,yang2025bidirectional}& Pixel/Voxel & \checkmark & $\times$ \\
     Points Representation\cite{guo2021unsupervised}& Landmark & \checkmark & $\times$ \\
     Shape Representation\cite{meng2023deepmesh,yuan20234d}& Mesh/SDF & $\times$ & $\times$ \\
    Dyna3DGR (ours)& 3D Gaussian & \checkmark &  \checkmark \\
    \bottomrule
    \end{tabular}
\end{table}

\section{Method}

\begin{figure}[t]
    \centering
    \includegraphics[width=\textwidth]{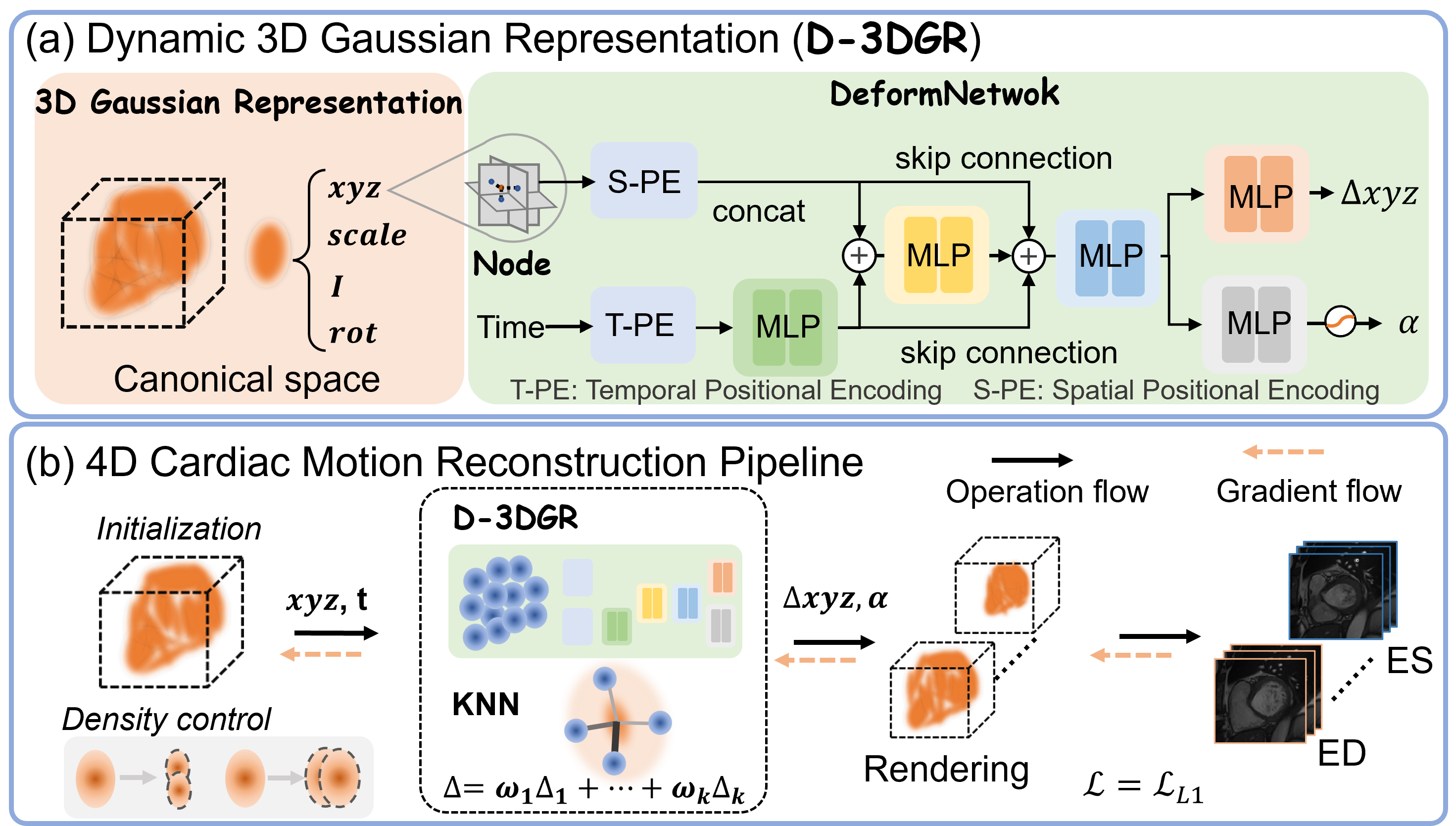}
    \caption{Overview of Dyna3DGR. (a) The Dyna3DGR consists of an explicit 3D Gaussian representation in canonical space (orange background) for volumetric reconstruction and an implicit motion representation powered by a deformation network (green background) for motion modeling. (b) The end-to-end pipeline that jointly optimizes both volumetric and motion representations.} 
    \label{fig:framework}
\end{figure}

In this section, we present the Dynamic 3D Gaussian Representation (Dyna3DGR) framework in Fig. \ref{fig:framework}. 
The framework integrates explicit 3D Gaussians for cardiac volume representation with the implicit neural field for motion field modeling.

\noindent \textbf{Explicit 3D Gaussian Representation:}
The 3D Gaussians are explicitly used to model cardiac volumetric structures. 
Following \cite{li2023sparse-gs,fu20243dgr}, each Gaussian $G_i$ has a parameter set $\phi_i = \{xyz_i, \Sigma_i, I_i\}$, where $xyz_i \in \mathbb{R}^3$ denotes the center position of the Gaussian and $\Sigma \in \mathbb{R}^{3\times3}$ is the covariance matrix. The $\Sigma$ can be decomposed into two learnable components: a quaternion $\boldsymbol{rot}$ and a scaling $\boldsymbol{scale}$. They can be transformed into the corresponding matrices $R$ and $S$. The resulting $\Sigma$ can be expressed as: $\Sigma = RSS^TR^T$.
The $I_i \in \mathbb{R}$ determines intensity value of Gaussian center.
Each Gaussian's influence to position $X$ is mathematically described by:
\begin{equation}
G_i(X|\phi_i) = I_i \cdot e^{-\frac{1}{2}(X-xyz_i)^T \Sigma_i^{-1}(X-xyz_i)},
\label{eq:3dgs}
\end{equation}
where $X \in \mathbb{R}^3$ denotes an position in the 3D space. The exponential term defines the spatial decay of the Gaussian's influence based on the distance from its center.
The volumetric value at any point $X$ is computed as a local aggregation of contributions from nearby Gaussians:
\begin{equation}
V(X|\theta_{{i}})=\sum_{i : ||X-xyz_i||\leq d_i}  G_i(X|\phi_i),
\end{equation}
where $d_i$ defines the effective radius of influence for each Gaussian. 

\noindent \textbf{Implicit Motion Representation:}
To effectively model cardiac tissue motion and deformation over time, we introduce a control-nodes-based deformation mechanism following \cite{huang2024sc-gs}. We employ a set of control nodes $C = \{C_i \in \mathbb{R}^3, o_i \in \mathbb{R}^+\}$,
where $C_i$ represents learnable position coordinates in the canonical space, $o_i$ is the learnable radius of a radial-basis-function kernel that controls the impact of a control node.
For each control node $C_i$, we learn a Gaussian transformation parameter $[\Delta xyz^t_i|\alpha^t_i]$ with a neural network. The transformation consists of a translation vector $\Delta xyz \in \mathbb{R}^3$ and a scale factor vector $\alpha \in \mathbb{R}^{+3}$. 
The deformation network $\mathcal{F_\theta}$ takes both the node position and temporal information as inputs to predict the transformation of these positions:
\begin{equation}
    (\Delta xyz, \alpha) = \mathcal{F_\theta}(\gamma(\operatorname{sg}(\boldsymbol{x})), \gamma(t)),
\end{equation} 
where $\operatorname{sg}(\cdot)$ indicates a stop-gradient operation, $\gamma$ denotes the positional encoding:
\begin{equation}
    \gamma(p) = (sin(2^k\pi p), cos(2^k \pi p))_{k=0}^{L-1}.
    \label{encoding}
\end{equation}

\noindent \textbf{DeformNetwork:}
The network takes an $N \times 3$ point cloud coordinate matrix and an $N \times 1$ temporal vector as input, outputting an $N \times 3$ translation vector $\Delta xyz$ and an $N \times 3$ scaling vector $\alpha$.
The framework employs the L1 loss function to guide the optimization process. The loss is computed between the predicted volumetric frames and the ground truth images:
\begin{equation}
\mathcal{L} = \sum||V(G(xyz+\Delta xyz, \alpha *scale,rot,I)) - V_{gt}||_1,
\end{equation}
where 
$V_{gt}$ denotes the ground truth cardiac voxels. 


To compute the dense motion field for each Gaussian in the canonical space, we employ a KNN search to identify the $k$ closest control points. The final transformation for each Gaussian is then derived through Linear Blend Skinning, which interpolates the transformations of the nearby control nodes to determine the position and scale changes of the Gaussian. This approach can be formulated as:
\begin{equation}
[\Delta xyz^t_i|\alpha^t_i] = \sum_{j=1}^k w_j \cdot [\Delta xyz^t_j|\alpha^t_j],
\end{equation}
where $w_j$ represents the blending weight calculated based on the distance between the Gaussian and the nearest $k$ control node. The blending weights are calculated by
\begin{equation}
\small
\label{eq:control_weight}
    w_{j} = \frac{\hat{w}_{ij}}{\sum\limits_{j \in \mathcal{C}_k} \hat{w}_{ij}} \text{, } \ \hat{w}_{ij} = \text{exp}(-\frac{d_{ij}^2}{2o_j^2}), 
\end{equation}
where $d_{ij}$ is the distance between center of Gaussian $G_i$ and the $j$-th neighboring control node, and $o_j$ is the learned radius parameter of the $j$-th control point.  



\section{Experiments}
\noindent \textbf{Dataset:}
Our method is evaluated on the Automated Cardiac Diagnosis Challenge (ACDC) dataset\cite{bernard2018deep}, a comprehensive collection of 4D cardiac MRI sequences. The dataset consists 100 clinical cases across five categories: normal subjects (NOR), myocardial infarction with systolic heart failure (MINF), dilated cardiomyopathy (DCM), hypertrophic cardiomyopathy (HCM), and abnormal right ventricle (ARV). Each case includes expert-annotated segmentation masks for three key cardiac structures in the end-diastole (ED) and end-systole (ES): the Left Ventricle (LV), Right Ventricle (RV) and Myocardium (Myo). 
\begin{table}[t]
    \centering
    \caption{Quantitative comparison of different methods. Tracking results evaluated with region-specific (RV,LV,Myo) and aggregate Dice score (\%), PSNR (dB) and SSIM (\%).(mean ± std, best and second-best results are in \textbf{bold} and \underline{underline}, respectively). }
    \label{tab: result1_part1}
    \begin{tabular}{l|cccccc}
    \toprule
    Method & 
    RV $\uparrow$ & 
    LV $\uparrow$ & 
    Myo $\uparrow$ & 
    Avg. $\uparrow$ & 
    PSNR $\uparrow$ & 
    SSIM $\uparrow$ 
    \\ \midrule
    \multicolumn{1}{l|}{LDDMM~\cite{beg2005computing}} & 73.61\tiny{$\pm$8.5} & 65.62\tiny{$\pm$8.5} & 56.44\tiny{$\pm$13} & 72.39\tiny{$\pm$18} & 31.20\tiny{$\pm$3.8} & \multicolumn{1}{c}{84.59\tiny{$\pm$6.0}} \\
    \multicolumn{1}{l|}{RDMM~\cite{shen2019region}} & 76.43\tiny{$\pm$7.8} & 69.50\tiny{$\pm$9.1} & 62.19\tiny{$\pm$14} & 75.51\tiny{$\pm$12} & 31.66\tiny{$\pm$3.9} & \multicolumn{1}{c}{84.36\tiny{$\pm$5.4}} \\
    \multicolumn{1}{l|}{ANTs (SyN)~\cite{avants2008symmetric}} & 75.30\tiny{$\pm$7.4} & 66.92\tiny{$\pm$8.6} & 58.03\tiny{$\pm$11} & 74.64\tiny{$\pm$13} & 30.92\tiny{$\pm$3.6} & \multicolumn{1}{c}{84.26\tiny{$\pm$5.6}} \\ \hline
    \multicolumn{1}{l|}{VoxelMorph~\cite{balakrishnan2019voxelmorph}} & 81.60\tiny{$\pm$6.5} & 77.00\tiny{$\pm$8.6} & 67.90\tiny{$\pm$13} & 79.90\tiny{$\pm$11} & 34.68\tiny{$\pm$3.3} & \multicolumn{1}{c}{85.01\tiny{$\pm$5.5}} \\
    \multicolumn{1}{l|}{DiffuseMorph~\cite{kim2022diffusemorph}} & 82.10\tiny{$\pm$6.7} & 78.30\tiny{$\pm$8.6} & 67.80\tiny{$\pm$15} & 80.50\tiny{$\pm$11} & 34.73\tiny{$\pm$3.6} & \multicolumn{1}{c}{84.30\tiny{$\pm$5.2}} \\
    \multicolumn{1}{l|}{CorrMLP~\cite{meng2024correlation}} & 80.33\tiny{$\pm$6.5} & 80.07\tiny{$\pm$7.8} & 70.51\tiny{$\pm$14} & 80.44\tiny{$\pm$8.6} & 34.90\tiny{$\pm$2.9} & \multicolumn{1}{c}{84.27\tiny{$\pm$4.5}} \\
    \multicolumn{1}{l|}{DeepTag~\cite{ye2021deeptag}} & 81.89\tiny{$\pm$7.0} & 79.10\tiny{$\pm$7.5} & 70.37\tiny{$\pm$13} & 80.83\tiny{$\pm$12} & 33.64\tiny{$\pm$3.4} & \multicolumn{1}{c}{83.09\tiny{$\pm$4.9}} \\
    \multicolumn{1}{l|}{Transmatch~\cite{chen2023transmatch}} &81.22\tiny{$\pm$7.0} &80.34\tiny{$\pm$6.8} &71.21\tiny{$\pm$12} &81.35\tiny{$\pm$9.8} &33.89\tiny{$\pm$3.3} & \multicolumn{1}{c}{84.78\tiny{$\pm$4.9}} \\
    \multicolumn{1}{l|}{FSDiffReg~\cite{qin2023fsdiffreg}} & 82.70\tiny{$\pm$6.1} & 80.90\tiny{$\pm$7.7} & 72.40\tiny{$\pm$12} & 82.30\tiny{$\pm$9.6} & 35.34\tiny{$\pm$3.5} & \multicolumn{1}{c}{85.85\tiny{$\pm$5.2}} \\ 
    \multicolumn{1}{l|}{GPTrack \cite{yang2025bidirectional}} & \underline{82.91\tiny{$\pm$5.8}} & \underline{81.23\tiny{$\pm$8.2}} & \underline{72.86\tiny{$\pm$9.0}} & \underline{82.65\tiny{$\pm$10}} & \textbf{35.52\tiny{$\pm$3.1}} & \multicolumn{1}{c} {\underline{86.19\tiny{$\pm$5.0}}} \\ \hline
    \multicolumn{1}{l|}{Dyna3DGR 
 (ours)} & \textbf{97.61\tiny{$\pm$1.4}} & \textbf{97.10\tiny{$\pm$2.3}} & \textbf{{95.16\tiny{$\pm$4.4}}} & \textbf{96.62\tiny{$\pm$2.7}} & \underline{34.66\tiny{$\pm$2.5}} & \multicolumn{1}{c}{\textbf{97.08\tiny{$\pm$0.25}}} \\
    \bottomrule
    \end{tabular}
\end{table}

\noindent \textbf{Implementation Details:}
Our 4D cardiac motion estimation framework employs an instance-wise test-time optimization strategy, requiring no pre-training or additional training data. We normalize spatial coordinates to the range [0,1]. The initial positions of 3D Gaussian primitives are obtained through uniform sampling of the ED phase segmentation mask positions, while control nodes are also initialized from these Gaussian positions (with the maximum number equal to the initial number of Gaussians).
The framework is implemented in PyTorch and optimized for 20,000 iterations (approximately 11 minutes). The optimization process follows a two-stage strategy: the first 1,000 iterations focus solely on optimizing the canonical space 3D Gaussians to establish stable positions and shapes, followed by joint optimization of 3D Gaussians, control node positions and the deformation network.
We employ the Adam optimizer \cite{KingBa15} with different initial learning rates for various components:
3D Gaussian positions: 1e-4;
Intensity values: 5e-3;
Rotation and scale parameters: 1e-4;
Control points: 1e-4 (update at iteration 5,000);
Deformation network: 1e-6.
The learning rates undergo exponential decay from 1e-4 to 1e-7, with optimizer parameters $\beta=(0.9, 0.999)$ and $\epsilon=1e-15$. Gaussian densification is performed every 500 iterations starting from the 500$^{th}$ iteration.
For the ACDC dataset preprocessing, we follow the state-of-the-art benchmark protocol presented in \cite{yang2025bidirectional}. All slices are resampled to 1.5×1.5×3.15mm resolution and center-cropped to 128×128×32 dimensions. The image intensities are normalized to [0,1].
All experiments were conducted on a single NVIDIA RTX 3090 GPU.

\begin{figure}[ht]
    \centering
    \includegraphics[width=\textwidth]{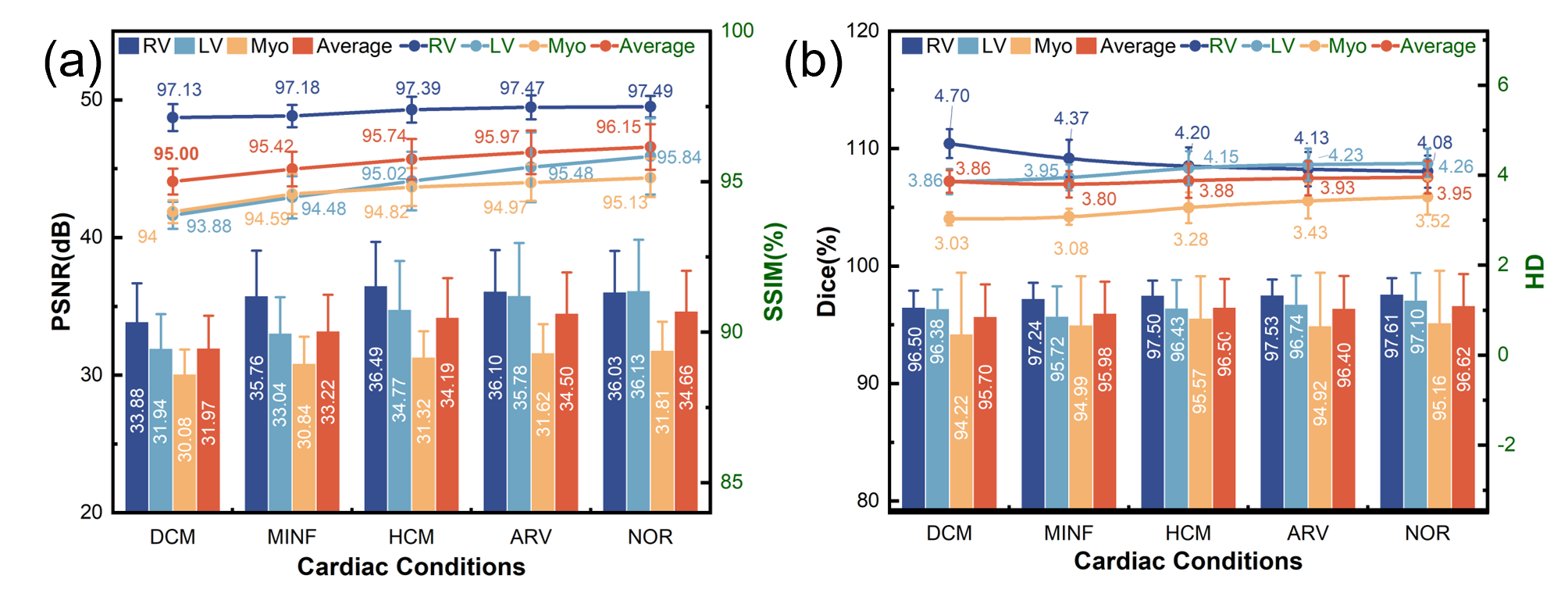}
    \caption{Quantitative comparison of the estimated ES frame of various cardiac condition.} 
    \label{fig:histogram}
\end{figure}


\noindent  \textbf{Evaluation Metrics:}
We employ multiple metrics to comprehensively evaluate our motion estimation framework. 
Following \cite{yang2025bidirectional}, 
the anatomical accuracy is assessed using the Dice score to measure the overlap between predicted cardiac segmentations and ground-truth annotations. Image reconstruction quality is evaluated through Peak Signal-to-Noise Ratio (PSNR) and Structural Similarity Index (SSIM) \cite{ssim} in Table \ref{tab: result1_part1}. 
The physical plausibility of predicted Gaussian position displacement field ($\Delta xyz$) is evaluated through diffeomorphic properties by examining the Jacobian determinant of the deformation field. Specifically, we count the number of locations where ($det(J_\phi)\leq 0\;$), which indicates points where the deformation exhibits undesirable properties such as folding or tearing that violate the topology-preserving requirement \cite{ye2021deeptag}. 
Following \cite{qin2023generative}, physiological plausibility is evaluated by the mean absolute deviation between Jacobian determinant of the predicted Gaussian position displacement field ($\Delta xyz$) and unity, denoted as $||J_\phi|-1|$, to measure the level of volume preservation, where significant deviations suggest violations of cardiac tissue incompressibility. The Haussdorff Distance (HD) results are also provided.
Model complexity is reported as the total number of parameters (in millions) for computational efficiency assessment.

\begin{table}[t]
    \centering
    \caption{Quantitative comparison of different methods. (mean ± std, best and second-best results are in \textbf{bold} and \underline{underline}, respectively). }
    \label{tab:result1_part2}
    \begin{tabular}{l|cccccc}
    \toprule
    Method & 
    HD $\downarrow$ &
    \footnotesize{$||J|-1|$} $\downarrow$ &
    \scriptsize{$det(J_\phi)\leq 0$} $\downarrow$ & 
    Params (M) $\downarrow$ 
    \\ \midrule
    \multicolumn{1}{l|}{LDDMM~\cite{beg2005computing}} & 6.562\tiny{$\pm$2.1} & 451.8\tiny{$\pm$162.3} & \multicolumn{1}{c}{653.5\tiny{$\pm$371.2}} & - \\
    \multicolumn{1}{l|}{RDMM~\cite{shen2019region}} & 5.728\tiny{$\pm$1.5} & 144.2\tiny{$\pm$63.67} & \multicolumn{1}{c}{266.0\tiny{$\pm$165.3}} & - \\
    \multicolumn{1}{l|}{ANTs (SyN)~\cite{avants2008symmetric}} & 6.242\tiny{$\pm$1.6} & 15.82\tiny{$\pm$22.30} & \multicolumn{1}{c}{57.26\tiny{$\pm$37.74}} & - \\ \hline
    \multicolumn{1}{l|}{VoxelMorph~\cite{balakrishnan2019voxelmorph}} & 5.336\tiny{$\pm$1.3} & 0.260\tiny{$\pm$0.070} & \multicolumn{1}{c}{0.079\tiny{$\pm$0.058}} & \underline{0.327} \\
    \multicolumn{1}{l|}{DiffuseMorph~\cite{kim2022diffusemorph}} & 3.977\tiny{$\pm$1.2} & 0.237\tiny{$\pm$0.068} & \multicolumn{1}{c}{0.061\tiny{$\pm$0.038}} & \underline{0.327} \\
    \multicolumn{1}{l|}{CorrMLP~\cite{meng2024correlation}} & 3.552\tiny{$\pm$1.3} & 0.248\tiny{$\pm$0.055} & \multicolumn{1}{c}{0.059\tiny{$\pm$0.022}} & 13.36 \\
    \multicolumn{1}{l|}{DeepTag~\cite{ye2021deeptag}} & 3.716\tiny{$\pm$1.4} & 0.185\tiny{$\pm$0.067} & \multicolumn{1}{c}{0.044\tiny{$\pm$0.025}} & 0.362 \\
    \multicolumn{1}{l|}{Transmatch~\cite{chen2023transmatch}} & 3.361\tiny{$\pm$1.1} & 0.226\tiny{$\pm$0.050} & 0.077\tiny{$\pm$0.054} & 70.71 \\
    \multicolumn{1}{l|}{FSDiffReg~\cite{qin2023fsdiffreg}} & \underline{3.283\tiny{$\pm$1.2}} & 0.214\tiny{$\pm$0.054} & \multicolumn{1}{c}{0.054\tiny{$\pm$0.026}} & 1.320 \\ 
    \multicolumn{1}{l|}{GPTrack \cite{yang2025bidirectional}} & \textbf{3.145\tiny{$\pm$1.1}} & \underline{0.178\tiny{$\pm$0.024}} & \multicolumn{1}{c} {\underline{0.032\tiny{$\pm$0.021}}} & 1.094 \\ \hline
    \multicolumn{1}{l|}{Dyna3DGR (ours)} & 3.940\tiny{$\pm$0.18} & \textbf{0.002\tiny{$\pm$0.0003}} & \multicolumn{1}{c}{\textbf{0.0\tiny{$\pm$0.0}}} & 0.604 \\
    \bottomrule
    \end{tabular}
    \label{tab: result1_part2} 
\end{table}

\noindent \textbf{Result:}
Our experiments demonstrate three key advantages of Dyna3DGR: (1) Superior anatomical accuracy with 17.73\% improvement in Dice score and 12.63\% increase in SSIM metrics compared to prior arts (paired t-test, p < 0.001); (2) Near-perfect physical plausibility (Jacobian deviation=0.002); (3) Lightweight architecture, fewer than most current methods.

\begin{figure}[t]
    \centering
    \includegraphics[width=0.95\textwidth]{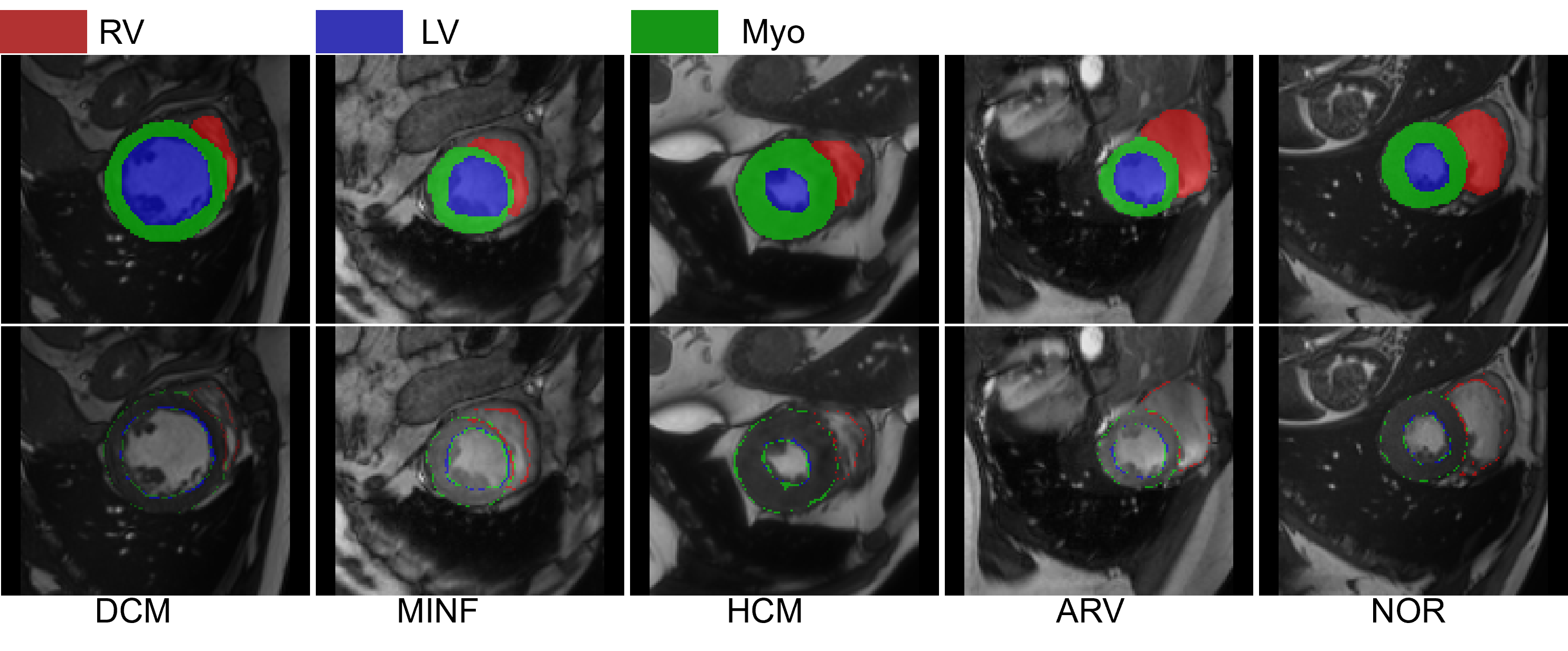}
    \caption{Visualization of tracking errors (second row) at ES frame and corresponding ground truth (first row) across varying cardiac conditions.} 
    \label{fig:error}
\end{figure}

The quantitative results presented in Tables \ref{tab: result1_part1} and \ref{tab: result1_part2} are evaluated following the same preprocessing protocol as GPTrack for fair comparison. Our method requires no training phase, and the reported metrics represent average performance of all samples. In terms of anatomical accuracy, Dyna3DGR consistently outperforms existing methods, maintaining above 90\% accuracy for both normal and pathological cases. 
Unlike discrete voxel representation, Gaussian ellipsoids influence intensities across a patch of voxels, and the orientation of the ellipsoids naturally encodes structural contour information. Therefore, learning deformations in Gaussian representation compared to learning voxel deformations better preserves contour information during motion, resulting in more accurate tracking and reconstruction of cardiac dynamics.

The results in Fig. \ref{fig:histogram} show Dyna3DGR's consistent tracking performance across various cardiac conditions, while Fig. \ref{fig:error} provides detailed visualization of tracking errors at the ES frame. 
The ablation study investigates the impact of varying the number of control nodes on model performance, with results presented in Table \ref{tab: Ablation}. The findings demonstrate a positive correlation between the number of control nodes and overall performance across all metrics.

\begin{table}[t]
    \centering
    \caption{Ablation study on numbers of control node. (mean ± std, best are in \textbf{bold})}
    \label{tab: Ablation}
    \begin{tabular}{l|lllll}
    \toprule
    Number & 2048 & 4096 & 6144 & 8192 & 10240  \\
    \midrule
    Dice (\%) $\uparrow$ & 92.51\tiny{$\pm$}2.34 & 94.60\tiny{$\pm$}1.78 & 95.27\tiny{$\pm$}1.99 & 95.70\tiny{$\pm$}2.30 & \textbf{96.62}\tiny{$\pm$}3.36 \\
    PSNR (dB) $\uparrow$ & 31.65\tiny{$\pm$}2.50 & 32.68\tiny{$\pm$}2.52 & 33.18\tiny{$\pm$}2.62 & 33.75\tiny{$\pm$}2.83 & \textbf{34.66}\tiny{$\pm$}2.43 \\
    SSIM (\%) $\uparrow$ & 94.78\tiny{$\pm$}0.35 & 95.99\tiny{$\pm$}0.32 & 96.40\tiny{$\pm$}0.30 & 96.72\tiny{$\pm$}0.27 & \textbf{97.08}\tiny{$\pm$}0.35 \\
    \footnotesize{$||J|-1|$} $\downarrow$ & 0.0041\tiny{$\pm$}0.001 & 0.0029\tiny{$\pm$}0.001 & 0.0030\tiny{$\pm$}0.001 & 0.0028\tiny{$\pm$}0.001 & \textbf{0.0021}\tiny{$\pm$}0.0003 \\
    \bottomrule
    \end{tabular}
\end{table}


\section{Conclusion}
This paper presents Dyna3DGR, that uniquely integrates explicit 3D Gaussian representation with implicit neural motion field modeling for cardiac motion estimation. The approach addresses key limitations of existing methods by eliminating the need for training data and bridging the gap between representation space and image space. Through self-supervised optimization, Dyna3DGR achieves accurate cardiac motion estimation. The evaluations on the ACDC dataset demonstrate that our method surpasses existing diffeomorphic registration approaches.

\begin{credits}
\subsubsection{\ackname} Supported by National Natural Science Foundation of China under Grant 62271465, Suzhou Basic Research Program under Grant SYG202338, and Jiangsu Province Science Foundation for Youths (NO. BK20240464).

\subsubsection{\discintname}
The authors have no competing interests to declare that are
relevant to the content of this article.
\end{credits}

\bibliographystyle{splncs04.bst}
\bibliography{reference}
\end{document}